\documentclass{article} 
\PassOptionsToPackage{table}{xcolor}
\usepackage{iclr2027_conference,times}

\usepackage{amsmath,amsfonts,bm}

\def\eqref#1{equation~\ref{#1}}

\def\1{\bm{1}}

\DeclareMathAlphabet{\mathsfit}{\encodingdefault}{\sfdefault}{m}{sl}
\SetMathAlphabet{\mathsfit}{bold}{\encodingdefault}{\sfdefault}{bx}{n}

\usepackage{amssymb,amsthm,mathrsfs,nicefrac}
\usepackage{microtype}

\usepackage{graphicx,float}
\usepackage{subcaption}
\usepackage{xcolor}
\definecolor{GUIFlowAccent}{RGB}{67,139,202}
\usepackage{array,tabularx,booktabs,multirow,makecell}
\usepackage{arydshln}

\usepackage[ruled]{algorithm2e}
\usepackage{soul,xspace}
\usepackage{url}
\usepackage{hyperref}
\hypersetup{colorlinks=true,linkcolor=GUIFlowAccent,citecolor=GUIFlowAccent,urlcolor=GUIFlowAccent}

\newcommand{\sysname}{GUIFlow\xspace}

\title{GUIFlow: Temporal Redundancy-Aware Incremental Inference for GUI Agents}

\author{
    \textbf{Ye Li}$^{1}$,
    \textbf{Haoran Ding}$^{1}$,
    \textbf{Huanan Liu}$^{1}$,
    \textbf{Yuan Meng}$^{1}$\thanks{\raggedright Corresponding Authors: yuanmeng@tsinghua.edu.cn, li-jw15@tsinghua.org.cn, wangzhi@sz.tsinghua.edu.cn.}\hphantom{$^{\ast}$},
    \textbf{Yutian Shen}$^{1}$, \\
    \textbf{Yuansong Wang}$^{1}$,
    \textbf{Kangye Ji}$^{1}$,
    \textbf{Hanyun Cui}$^{1}$,
    \textbf{Hongbo Duan}$^{1}$,
    \textbf{Chen Tang}$^{2}$, \\
    \textbf{Jiawei Li}$^{3\ast}$,
    \textbf{Shu-Tao Xia}$^{1}$,
    \textbf{Zhi Wang}$^{1\ast}$ \\
    $^1$ Tsinghua University \quad
    $^2$ The Chinese University of Hong Kong \\
    $^3$ Joy Future Academy, JD Group, Beijing, China
}

\iclrfinalcopy
\begin{document}

\maketitle

\begin{abstract}
GUI agents built on vision-language models interact with continuously evolving interfaces, yet conventional inference repeatedly recomputes representations of largely overlapping observations and histories. This mismatch makes multi-step interaction costly, while changes in context, interface content, and action arguments prevent straightforward cache reuse. We introduce GUIFlow, a framework for full-pipeline incremental inference in multi-image GUI interaction. Our central insight is that advancing a decision need not require rebuilding all supporting representations: each interaction step can instead selectively update computational states maintained across the trajectory. To support new decisions as the context evolves, GUIFlow adapts how past computation is reused: historical KV states are position-aligned and injected into the current context; change signals guide selective updates of current-observation states in the vision encoder and language model; and earlier action responses serve as speculative drafts verified by the target model. Together, these mechanisms enable cross-step reuse throughout visual encoding, prefill, and decoding without additional training, a separate draft model, or changes to the model-visible inputs and action interface. Experiments demonstrate up to $3.39\times$ geometric-mean speedup across offline multi-image benchmarks and up to $3.49\times$ in online, closed-loop interaction on OSWorld, while maintaining task performance broadly comparable to the dense baseline.
\end{abstract}

\section{Introduction}
\label{sec:introduction}

AI agents are moving beyond answering questions toward executing tasks in digital environments, including everyday applications on computers and mobile devices.
Graphical user interfaces (GUIs) provide a broadly applicable way to interact with existing software without relying solely on application-specific APIs.
Advances in vision-language models~\citep{bai2025qwen3} have enabled GUI agents to interpret screens and act on them through repeated observation and action~\citep{qin2025ui,xu2026mobile,xie2024osworld}.
However, completing multi-step tasks requires sustained model inference, whose cumulative latency limits the responsiveness and practical usability of these agents.

The continuity of interaction offers an opportunity to reduce this cost.
Successive decision steps share supporting information: historical screenshots recur across overlapping context windows, neighboring observations share persistent interface regions, and action responses often repeat structural patterns.
For example, editing a text field changes only part of a screen, while the surrounding interface and much of the interaction history remain available from the preceding step.
Figure~\ref{fig:temporal_redundancy} provides evidence at both the image and representation levels: across the evaluated tasks, more than $70\%$ of observation pairs have RGB-image error no greater than $5\%$ at each temporal gap from one to four steps.
The illustrated KV-drift maps show similar spatial patterns of change in an early and the final layer, while the trajectory example shows substantial state stability between major interface changes.
These observations suggest that advancing a decision can build on persistent computational states and selectively update what changes, rather than rebuild all supporting representations at every step.

Recent efficiency methods exploit temporal structure through KV-cache compression, as in GUI-KV and ST-Lite~\citep{huang2025gui,zhou2026efficient}, or learned visual-patch reduction, as in ReVision~\citep{abaskohi2026revision}.
Reducing visual context, however, does not by itself make past computational states reusable under new conditions or ensure benefits across the full inference pipeline, raising two broader challenges.
The first is \textbf{maintaining reusable computation under evolving interactions}: past computational states must support new decisions even as their context changes, requiring a distinction between what remains reusable and what needs updating.
The second is \textbf{turning local reuse into pipeline-wide acceleration}: savings within individual stages may be limited by computation elsewhere and by the overhead of reuse itself.
Addressing this challenge requires extending reuse across visual encoding, prefill, and decoding while translating computational savings into actual latency reductions.
These challenges motivate our central question: \emph{Can GUI agents accelerate real-world interaction through full-pipeline incremental inference, selectively updating existing computational states as the interaction evolves?}

To address these challenges, we introduce \textbf{GUIFlow}, a temporal redundancy-aware framework for full-pipeline incremental inference in GUI agents.
GUIFlow treats each interaction step as a selective update to computational states maintained across the trajectory, rather than an independent inference pass.
For historical inputs, it injects position-aligned KV states for recurring screenshots into the current context, enabling approximate reuse beyond unchanged prefixes.
For current observations, change signals guide selective updates in the vision encoder and deeper language-model layers, while source tracking and reuse limits control state staleness.
For action generation, earlier responses serve as speculative drafts rather than fixed answers, with the target model verifying draft tokens and correcting mismatches.
Together, these mechanisms extend reuse across visual encoding, prefill, and decoding, reducing repeated computation beyond any single stage.
GUIFlow requires neither additional training nor a separate draft model, and leaves the model-visible inputs and action interface unchanged.

We evaluate GUIFlow on AITW~\citep{rawles2023androidinthewild} and Mind2Web~\citep{deng2023mind2web,cheng2024seeclick} for offline multi-image navigation and on OSWorld~\citep{xie2024osworld} for online closed-loop interaction, covering both fixed trajectories and evolving environments.
GUIFlow achieves up to $3.39\times$ geometric-mean speedup across the offline benchmarks and up to $3.49\times$ speedup on OSWorld, while maintaining task performance broadly comparable to the dense baseline.
Across these settings, mean per-step processing latency is reduced by factors of $1.94$--$2.55$.
These results show that selective state updates can turn temporal redundancy into both computational savings and lower execution latency. Our contributions are threefold:

\begingroup
\settowidth{\labelwidth}{(3)}
\setlength{\leftmargini}{\labelwidth}
\addtolength{\leftmargini}{\labelsep}
\begin{itemize}
\item [(1)] We introduce an incremental inference paradigm for GUI agents.
This paradigm reframes successive decisions as selective updates to persistent computational states rather than independent inference passes, exploiting temporal redundancy across interaction steps.
\item [(2)] We propose GUIFlow, a training-free framework for full-pipeline incremental inference.
GUIFlow integrates historical KV injection, source-aware current-state refinement, and verified action generation to extend cross-step reuse across visual encoding, prefill, and decoding.
\item [(3)] We demonstrate the effectiveness of GUIFlow in both offline navigation and online closed-loop interaction.
Experiments on AITW, Mind2Web, and OSWorld show substantial speedups and lower processing latency while maintaining task performance broadly comparable to dense inference.
\end{itemize}
\endgroup

\section{Related Work}
\label{sec:related_work}

\textbf{Efficient GUI Agents.}
Visual-token reduction~\citep{alvar2025divprune,li2025prance}, operator skipping~\citep{luo2026attend}, and KV-cache compression~\citep{li2024ems} offer complementary routes to reducing inference cost.
FastV prunes visual tokens after early decoder layers~\citep{chen2024image}, whereas VisionZip selects dominant tokens and merges contextual information before language-model processing~\citep{yang2025visionzip}.
For GUI interaction, ShowUI introduces interface-aware token selection~\citep{lin2025showui}.
GUI-KV, STaR-KV, and ST-Lite exploit spatial importance and temporal structure to compress screenshot-history KV caches~\citep{huang2025gui,han2026star,zhou2026efficient}, while ReVision learns to remove temporally redundant visual patches and fine-tunes the agent on compressed histories~\citep{abaskohi2026revision}.
Complementing input and cache compression, GUIFlow focuses on selectively updating persistent computational states across successive decisions.

\textbf{Computation Reuse and Incremental Inference.}
Temporal redundancy motivates both state caching~\citep{xu2026vla,ji2026blockwise} and temporal-aware acceleration~\citep{li2025sp,li2026elegantvla,li2025ts} in embodied control.
For language models, prefix caching reuses KV states for identical leading contexts, but recurring content outside an unchanged prefix requires additional treatment of positional and contextual changes.
CacheBlend addresses this issue in retrieval-augmented generation by selectively recomputing states to account for cross-chunk interactions~\citep{yao2025cacheblend}.
TRACE combines visual-token selection with cross-step cache reuse for GUI agents~\citep{wang2026trace}.
At the generation stage, speculative decoding evaluates candidate continuations with the target model, allowing multiple tokens to be accepted in one verification pass~\citep{leviathan2023fast}; REST retrieves drafts from existing text without a separate draft model~\citep{he2024rest}.
GUIFlow coordinates historical-state reuse, current-observation refinement, and draft verification under evolving contexts, organizing visual encoding, prefill, and decoding into a unified incremental inference process.

\section{Method}
\label{sec:method}

\begin{figure}[!t]
  \centering
  \includegraphics[width=\linewidth]{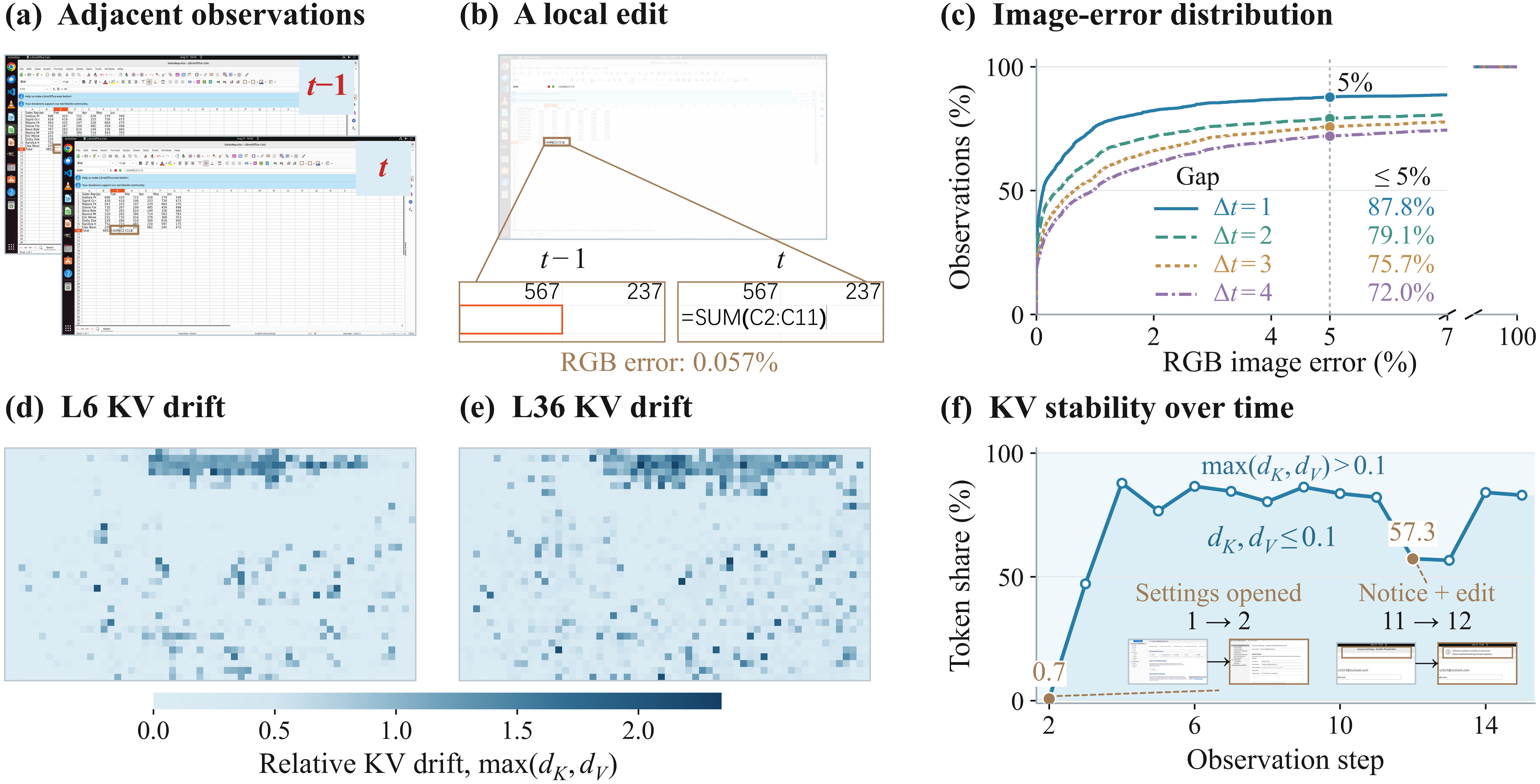}
  \caption{
    \textbf{Incremental changes in GUI observations and internal representations.}
    \textbf{(a--c) Image-level changes.} A local edit modifies only a small part of successive screenshots; across the evaluated tasks, more than $70\%$ of observation pairs have RGB-image error at most $5\%$ at each temporal gap $\Delta t\in\{1,2,3,4\}$.
    \textbf{(d--f) Representation-level changes.} For the same observation pair, KV-drift maps at layer 6 and the final layer 36 show broadly similar spatial concentrations of change.
    Along the illustrated trajectory, a large share of tokens has small KV drift ($d_K,d_V\leq0.1$) during stable-interface segments, with drops around substantial interface changes.
    Together, these observations motivate selectively updating changing states rather than rebuilding all representations at every step.
  }
  \label{fig:temporal_redundancy}
\end{figure}

\begin{figure}[!t]
  \centering
  \includegraphics[width=\linewidth]{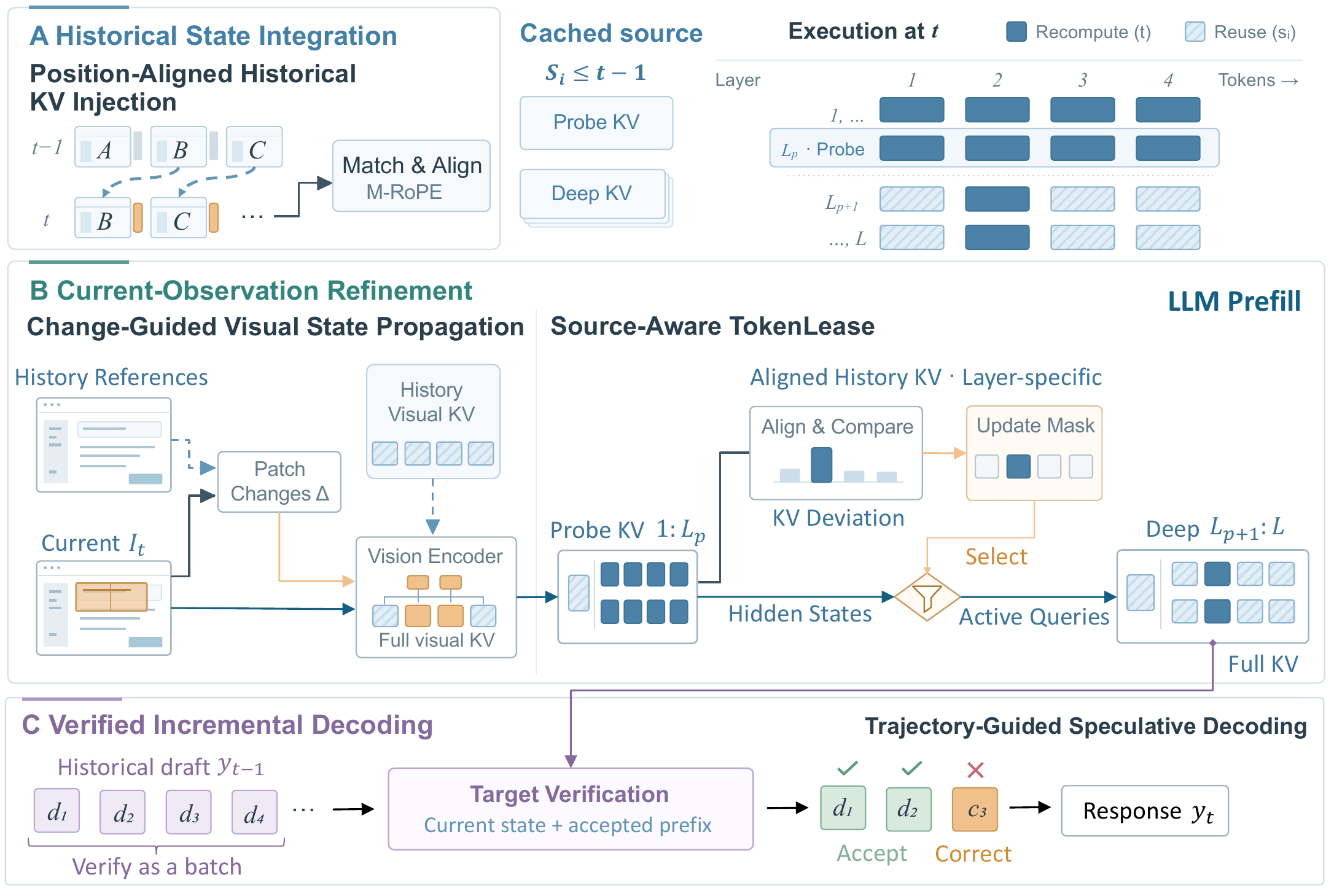}
  \caption{
    \textbf{Overview of \sysname.}
    \sysname treats successive GUI decisions as incremental updates to existing computation rather than independent inference passes.
    \textbf{(A) Historical State Integration.} Historical KV states are positionally aligned and injected into the current context.
    \textbf{(B) Current-Observation Refinement.} Pixel changes guide visual-encoder updates, while source-aware KV deviations guide deeper language-model updates.
    \textbf{(C) Verified Incremental Decoding.} Conditioned on the prefill state formed by (A) and (B), the target model verifies historical response drafts, accepting matching tokens and correcting mismatches.
    Together, these mechanisms exploit temporal redundancy across visual encoding, prefill, and decoding without additional training or a separate draft model.
  }
  \label{fig:guiflow_framework}
\end{figure}

\subsection{Full-Pipeline Incremental Inference}
\label{sec:incremental_framework}

At step $t$, a task instruction $q$, current screenshot $I_t$, and history $H_t$ of preceding screenshots and actions form the agent's input $X_t$.
Step-independent execution encodes the screenshots, prefills the language model, and autoregressively generates a response $y_t$ for action execution.
Despite KV caching within each response, this execution can repeatedly process overlapping histories and persistent interface content across steps.
Figure~\ref{fig:temporal_redundancy} examines whether changes between decisions require rebuilding these representations.
At the image level, panels (a--b) illustrate a localized edit, while panel (c) aggregates observation pairs across tasks: the share with RGB-image error at most $5\%$ is $87.8\%$ for adjacent steps and remains $72.0\%$ at a four-step gap.
Thus, visual continuity persists beyond immediately adjacent observations, providing opportunities to carry forward existing visual states.

At the representation level, panels (d--e) show that the same observation pair produces broadly similar spatial concentrations of KV drift at layer 6 and the final layer 36.
Panel (f) further shows that a large share of tokens retains small KV drift during stable-interface segments, whereas substantial interface changes coincide with drops in this share.
These patterns motivate using early-layer deviations to guide deeper-state updates and monitoring state changes as interaction progresses, rather than treating every new screenshot as entirely new computation.
Because position and preceding context also affect KV states, visual similarity alone is not sufficient to decide reuse.

Building on these observations, GUIFlow treats each new decision as a selective update to computation maintained within the same task trajectory.
Let $\mathcal{C}_{t-1}$ denote the state retained before step $t$.
Incremental inference produces both the current response and the state available to the next step:
\begin{equation}
    (y_t, \mathcal{C}_t)
    = \mathcal{G}^{\mathrm{inc}}_{\theta}(X_t, \mathcal{C}_{t-1}),
    \label{eq:incremental_inference}
\end{equation}
where $\mathcal{G}^{\mathrm{inc}}_{\theta}$ is an execution procedure for the fixed model with parameters $\theta$, not a separately trained model.
The maintained state includes historical KV caches, visual and language-layer caches with reuse metadata, and previous responses used as drafts.
Cached representations replace selected computations, whereas response drafts require target-model verification before acceptance.
The updated state $\mathcal{C}_t$ retains reusable computation and incorporates newly computed states and the current response.

GUIFlow organizes incremental execution around three sources of temporal redundancy, as illustrated in Figure~\ref{fig:guiflow_framework}.
First, historical screenshots recur across overlapping context windows, motivating reuse of their cached KV states with positional alignment as the window advances.
Second, successive observations share persistent interface regions, motivating selective refresh of visual and language states based on observed changes and representation deviations.
Finally, action responses within a trajectory often share structural patterns and token sequences, motivating historical drafts whose candidate tokens are verified by the target model and corrected at mismatches.
Unavailable or incompatible caches trigger initialization or recomputation.
Together, these mechanisms extend incremental execution across visual encoding, prefill, and decoding without changing model-visible inputs or the action interface.
Intermediate-state reuse can be approximate, and draft verification under the resulting prefill state does not imply equivalence to dense inference.

\subsection{Incremental State Integration for Historical Context}
\label{sec:historical_reuse}

We first address redundant computation in historical context, focusing on the KV states of screenshots retained across interaction steps.
As the context window advances, these screenshots may shift in position, requiring a mechanism beyond unchanged-prefix matching.

\textbf{Position-Aligned Historical KV Injection.}
The principle is to populate the current context with position-aligned historical KV states and compute the remaining regions, without adding tokens or changing the model-visible input.
GUIFlow maps eligible historical image blocks in $X_t$ to their cached counterparts in $\mathcal{C}_{t-1}$.
It first identifies a contiguous matching prefix using multimodal fingerprints: text tokens are identified by token ID and type, while image tokens are identified by image content, grid layout, and within-image offset.
Beyond this prefix, historical image blocks are matched by content, grid, and token count.
This mapping separates blocks eligible for cache injection from regions requiring computation, including intervening text and action tokens outside the matching prefix and historical images that cannot be reused.

For each matched historical token, GUIFlow transfers the cached KV states to the corresponding current position.
Let $p_s$ and $p_t$ denote its source and current multimodal positions, and let $R(p)$ denote the corresponding M-RoPE rotation.
At layer $\ell$, the position-aligned states are
\begin{equation}
    \widetilde{K}^{(\ell)}_t
    = R(p_t)R(p_s)^{\top}K^{(\ell)}_s,
    \qquad
    \widetilde{V}^{(\ell)}_t = V^{(\ell)}_s,
    \label{eq:historical_position_alignment}
\end{equation}
where $K^{(\ell)}_s$ already carries the source positional rotation; the correction acts on rotary dimensions and leaves other dimensions unchanged.
These states populate their slots in the current logical KV layout, allowing recomputed queries to attend to both reused and newly computed states under the current causal mask.

Finally, GUIFlow uses a positive integer hyperparameter $M_{\mathrm{hist}}$ to cap the number of sliding reuses of historical-image KV states, preventing indefinite reuse without recomputation.
Each sliding copy increments the inherited per-token reuse count, whereas exact-prefix reuse leaves it unchanged.
Before a historical image block is reused, its maximum source-token count is checked against $M_{\mathrm{hist}}$; if the count is at least $M_{\mathrm{hist}}$, the block is recomputed and its counts are reset to zero.

\subsection{Incremental State Refinement for Current Observations}
\label{sec:current_refresh}

The current observation captures the latest environment state and directly informs the next action, while sharing persistent interface regions and potentially reusable internal states with preceding observations.
GUIFlow therefore selectively refreshes current-observation computation at two levels, using pixel changes to guide visual-state updates and context-dependent representation deviations to guide deeper language-model updates.

\textbf{Change-Guided Visual State Propagation.}
GUIFlow propagates existing visual states while concentrating new computation on changed regions.
Patch-level pixel differences against a rolling reference identify an update set $\mathcal{A}_t$, expanded to include complete patch-merging groups.
At vision-encoder layer $\ell$, the KV states are updated as
\begin{equation}
    Z_{t,i}^{(\ell)} =
    \begin{cases}
        \widehat{Z}_{t,i}^{(\ell)}, & i\in\mathcal{A}_t,\\
        Z_{\mathrm{ref},i}^{(\ell)}, & i\notin\mathcal{A}_t,
    \end{cases}
    \qquad Z\in\{K,V\},
    \label{eq:visual_state_propagation}
\end{equation}
where $\widehat{Z}_{t,i}^{(\ell)}$ denotes a newly computed state and $Z_{\mathrm{ref},i}^{(\ell)}$ its cached counterpart in the rolling reference.
Selected patch queries attend to this full mixture of new and cached visual KV and undergo feed-forward computation; unselected patches retain their cached states.
Updated states become the next reference; incompatible layouts, extensive changes, or reaching the sparse-update limit trigger full recomputation.

\textbf{Source-Aware TokenLease.}
TokenLease probes context-dependent representation changes before deciding which deep states to refresh.
Current-image tokens pass through the first $L_p$ language-model layers, and their probe KV is compared with the position-aligned probe KV of their actual deep-cache sources, not necessarily the previous step.
We quantify this deviation and its residual mass as
\begin{equation}
    d_i = \max_{Z\in\{K,V\}}
    \frac{\operatorname{RMS}(Z_i^{\mathrm{cur}}-\bar{Z}_i^{\mathrm{src}})}
    {\max\bigl(\operatorname{RMS}(\bar{Z}_i^{\mathrm{src}}),\delta\bigr)},
    \qquad m_i=d_i^2,
    \label{eq:source_aware_kv_deviation}
\end{equation}
where $Z_i^{\mathrm{cur}}$ and $\bar{Z}_i^{\mathrm{src}}$ denote the current and position-aligned source probe states at layer $L_p$, respectively, and $\delta>0$ stabilizes the normalization.
The refresh set $\mathcal{R}_t$ includes pixel-changed tokens and tokens reaching the reuse limit $M_{\mathrm{cur}}$, then expands in descending residual mass until
\begin{equation}
    \sum_{i\in\mathcal{I}_t\setminus\mathcal{R}_t} m_i
    \leq \varepsilon \sum_{i\in\mathcal{I}_t} m_i,
    \label{eq:current_residual_budget}
\end{equation}
where $\mathcal{I}_t$ indexes current-image tokens and $\varepsilon\in[0,1]$ sets the unrefreshed residual-mass budget.
Beyond layer $L_p$, selected image queries are computed and others reuse deep KV; text queries and the full logical KV layout are retained.
Refreshed tokens renew their sources and reset reuse counts; reused tokens retain their sources and increment counts.

\subsection{Verified Incremental Decoding for Action Generation}
\label{sec:incremental_decoding}

We next accelerate autoregressive decoding by exploiting temporal redundancy in action responses.
Outputs at successive interaction steps often share structural patterns and token sequences, although action arguments may change with the current observation.
Rather than regenerating these sequences token by token, GUIFlow uses historical outputs to propose continuations and verifies them under the current context.

\begin{table}[!t]
  \centering
  \caption{
    \textbf{Main results on offline multi-image navigation.}
    FLOPs (T) are averaged per step; speedup is the Dense/method FLOPs ratio, with an equal-weight geometric mean across datasets.
    \sysname achieves $3.37$--$3.39\times$ geometric-mean speedups while maintaining scores close to Dense.
  }
  \label{tab:offline_navigation}
  \begingroup
  \footnotesize
  \setlength{\tabcolsep}{2pt}
  \renewcommand{\arraystretch}{1.12}
  \begin{tabularx}{\linewidth}{@{}l l *{7}{>{\centering\arraybackslash}X}}
    \toprule
    & & \multicolumn{3}{c}{\textbf{AITW}} & \multicolumn{3}{c}{\textbf{Mind2Web}} & \\
    \cmidrule(lr){3-5}\cmidrule(lr){6-8}
    \textbf{Model} & \textbf{Method}
    & \makecell{Action\\Matching\\(\%) $\uparrow$}
    & \makecell{FLOPs\\(T) $\downarrow$}
    & \makecell{Speedup\\($\times$) $\uparrow$}
    & \makecell{Step SR\\(\%) $\uparrow$}
    & \makecell{FLOPs\\(T) $\downarrow$}
    & \makecell{Speedup\\($\times$) $\uparrow$}
    & \makecell{\textbf{Geo.}\\\textbf{Mean}\\($\times$) $\uparrow$} \\
    \midrule
    \multirow{8}{*}{\makecell[l]{\textbf{Qwen3-VL}\\\textbf{4B-Instruct}}}
    & Dense & 61.20 & 31.94 & 1.00 & 49.77 & 43.97 & 1.00 & 1.00 \\
    & FastV & 60.26 & 22.14 & 1.44 & 48.31 & 30.10 & 1.46 & 1.45 \\
    & VisionZip & 60.23 & 18.91 & 1.69 & 42.09 & 24.90 & 1.77 & 1.73 \\
    & GUI-KV & 58.59 & 31.94 & 1.00 & 49.25 & 43.96 & 1.00 & 1.00 \\
    & STaR-KV & 60.34 & 31.93 & 1.00 & 49.61 & 43.95 & 1.00 & 1.00 \\
    & Op-Skip & 58.29 & 24.65 & 1.30 & \textbf{53.41} & 34.21 & 1.29 & 1.29 \\
    & HistPrune & 61.20 & 25.47 & 1.25 & 49.75 & 34.67 & 1.27 & 1.26 \\
    & \cellcolor{GUIFlowAccent!15}\sysname
    & \cellcolor{GUIFlowAccent!15}\textbf{61.74}
    & \cellcolor{GUIFlowAccent!15}\textbf{9.23}
    & \cellcolor{GUIFlowAccent!15}\textbf{3.46}
    & \cellcolor{GUIFlowAccent!15}49.16
    & \cellcolor{GUIFlowAccent!15}\textbf{13.37}
    & \cellcolor{GUIFlowAccent!15}\textbf{3.29}
    & \cellcolor{GUIFlowAccent!15}\textbf{3.37} \\
    \midrule
    \multirow{8}{*}{\makecell[l]{\textbf{UI-TARS}\\\textbf{1.5-7B}}}
    & Dense & 69.06 & 62.45 & 1.00 & \textbf{53.40} & 85.78 & 1.00 & 1.00 \\
    & FastV & 67.99 & 41.27 & 1.51 & 52.40 & 55.72 & 1.54 & 1.53 \\
    & VisionZip & 66.11 & 42.01 & 1.49 & 49.31 & 57.54 & 1.49 & 1.49 \\
    & GUI-KV & \textbf{69.12} & 62.45 & 1.00 & 52.17 & 85.78 & 1.00 & 1.00 \\
    & STaR-KV & 69.10 & 62.44 & 1.00 & 53.05 & 85.77 & 1.00 & 1.00 \\
    & Op-Skip & 65.17 & 50.30 & 1.24 & 49.49 & 69.20 & 1.24 & 1.24 \\
    & HistPrune & 69.06 & 52.30 & 1.19 & 53.29 & 71.58 & 1.20 & 1.20 \\
    & \cellcolor{GUIFlowAccent!15}\sysname
    & \cellcolor{GUIFlowAccent!15}68.93
    & \cellcolor{GUIFlowAccent!15}\textbf{18.33}
    & \cellcolor{GUIFlowAccent!15}\textbf{3.41}
    & \cellcolor{GUIFlowAccent!15}53.13
    & \cellcolor{GUIFlowAccent!15}\textbf{25.49}
    & \cellcolor{GUIFlowAccent!15}\textbf{3.37}
    & \cellcolor{GUIFlowAccent!15}\textbf{3.39} \\
    \bottomrule
  \end{tabularx}
  \endgroup
\end{table}

\textbf{Trajectory-Guided Speculative Decoding.}
Following the draft--verify principle of speculative decoding~\citep{leviathan2023fast}, GUIFlow treats the action-response portion of an earlier output, such as $y_{t-1}$, as a draft for $y_t$.
The candidate pool retains the most recent response for each action type within the same task, allowing earlier steps to supply alternatives when the immediately preceding action differs.
These drafts require neither a separate model nor additional training and are not appended to the model-visible prompt as extra history.
Instead, their candidate tokens are evaluated in a batched target-model forward pass from the current prefill state and the accepted response prefix.

For greedy decoding, let $u$ denote the accepted prefix and $d^{(j)}$ a candidate continuation of length $n_j$.
Writing $p_{\theta,t}$ for the target-model distribution conditioned on the current prefill state, we compute
\begin{equation}
    \begin{aligned}
        \widehat{d}^{(j)}_i &= \arg\max_v p_{\theta,t}(v\mid u,d^{(j)}_{<i}),\\
        a_j &= \max\bigl\{k\in\{0,\ldots,n_j\}:\ d^{(j)}_i=\widehat{d}^{(j)}_i\ \text{for all }1\leq i\leq k\bigr\}.
    \end{aligned}
    \label{eq:draft_prefix_acceptance}
\end{equation}
GUIFlow selects the longest matching prefix, favoring more recent drafts in a tie.
It retains the accepted tokens and their newly computed KV, discards rejected candidate states, and generates a target-model correction at the first mismatch before continuing decoding.
When applicable, later structural matches allow re-entry into a historical draft; otherwise, generation proceeds token by token.
Matching drafts thus let one verification pass advance multiple output tokens, reducing serial decoding calls while leaving the current token decisions to the target model.

\section{Experiments}
\label{sec:experiments}

\begin{table}[!t]
  \centering
  \caption{
    \textbf{Main results on online OSWorld interaction.}
    FLOPs (T) are averaged per step/request for each evaluated run; speedups are Dense/method mean FLOPs ratios across rollout trajectories.
    \sysname achieves $3.32$--$3.49\times$ speedups relative to Dense.
  }
  \label{tab:online_osworld}
  \begingroup
  \footnotesize
  \setlength{\tabcolsep}{4pt}
  \renewcommand{\arraystretch}{1.12}
  \begin{tabularx}{\linewidth}{@{}l l *{4}{>{\centering\arraybackslash}X}}
    \toprule
    \textbf{Model} & \textbf{Method}
    & \makecell{\textbf{Mean Score}\\(\%) $\uparrow$}
    & \makecell{\textbf{Strict SR}\\(\%) $\uparrow$}
    & \makecell{\textbf{FLOPs}\\(T) $\downarrow$}
    & \makecell{\textbf{Speedup}\\($\times$) $\uparrow$} \\
    \midrule
    \multirow{7}{*}{\makecell[l]{\textbf{Qwen3-VL}\\\textbf{8B-Instruct}}}
    & Dense & 24.30 & 23.55 & 247.01 & 1.00 \\
    & FastV & 23.16 & 22.16 & 165.50 & 1.49 \\
    & VisionZip & 21.22 & 20.22 & 161.48 & 1.53 \\
    & GUI-KV & 22.81 & 21.61 & 246.54 & 1.00 \\
    & HistPrune & 20.82 & 19.94 & 194.52 & 1.27 \\
    & DivPrune & 22.93 & 22.71 & 185.89 & 1.33 \\
    & \cellcolor{GUIFlowAccent!15}\sysname
    & \cellcolor{GUIFlowAccent!15}\textbf{27.90}
    & \cellcolor{GUIFlowAccent!15}\textbf{26.87}
    & \cellcolor{GUIFlowAccent!15}\textbf{70.85}
    & \cellcolor{GUIFlowAccent!15}\textbf{3.49} \\
    \midrule
    \multirow{6}{*}{\makecell[l]{\textbf{GUI-Owl}\\\textbf{1.5-8B-Instruct}}}
    & Dense & \textbf{48.79} & \textbf{47.09} & 278.12 & 1.00 \\
    & FastV & 45.83 & 44.32 & 187.36 & 1.48 \\
    & VisionZip & 44.36 & 42.38 & 183.18 & 1.52 \\
    & HistPrune & 45.74 & 43.77 & 219.07 & 1.27 \\
    & DivPrune & 41.79 & 40.17 & 182.84 & 1.52 \\
    & \cellcolor{GUIFlowAccent!15}\sysname
    & \cellcolor{GUIFlowAccent!15}46.30
    & \cellcolor{GUIFlowAccent!15}44.60
    & \cellcolor{GUIFlowAccent!15}\textbf{83.81}
    & \cellcolor{GUIFlowAccent!15}\textbf{3.32} \\
    \bottomrule
  \end{tabularx}
  \endgroup
\end{table}

\subsection{Main Results}
\label{sec:main_results}

\textbf{Offline Multi-Image Navigation.}
Many multimodal baselines optimize individual requests, while GUI-specific alternatives exploit temporal structure through context compression or historical caching.
Repeated screenshot histories and locally changing interfaces offer a broader opportunity: advancing a decision need not require recomputing all supporting representations.
Table~\ref{tab:offline_navigation} shows that \sysname achieves the lowest FLOPs across both models and datasets, with geometric-mean speedups of $3.37\times$ on Qwen3-VL and $3.39\times$ on UI-TARS while maintaining task scores close to Dense.
Across the four model--dataset settings, \sysname delivers $3.29$--$3.46\times$ speedups, compared with at most $1.77\times$ among the evaluated baselines.

\textbf{Online Closed-Loop Interaction.}
We complement offline action prediction on fixed trajectories with closed-loop evaluation in OSWorld, where actions alter subsequent observations and task success is evaluated from the resulting environment state.
This coupling requires cached computation to support decisions as the trajectory evolves.
\sysname achieves $3.49\times$ and $3.32\times$ speedups on Qwen3-VL and GUI-Owl, respectively, with higher mean scores and strict success rates than the evaluated acceleration baselines (Table~\ref{tab:online_osworld}).
On Qwen3-VL, \sysname reaches $27.90\%$ mean score and $26.87\%$ strict success rate, compared with $24.30\%$ and $23.55\%$ for Dense.
These findings extend the evidence for incremental inference from fixed histories to interactive execution; we examine wall-clock benefits in the following analysis.

\begin{figure}[!t]
  \centering
  \includegraphics[width=\linewidth]{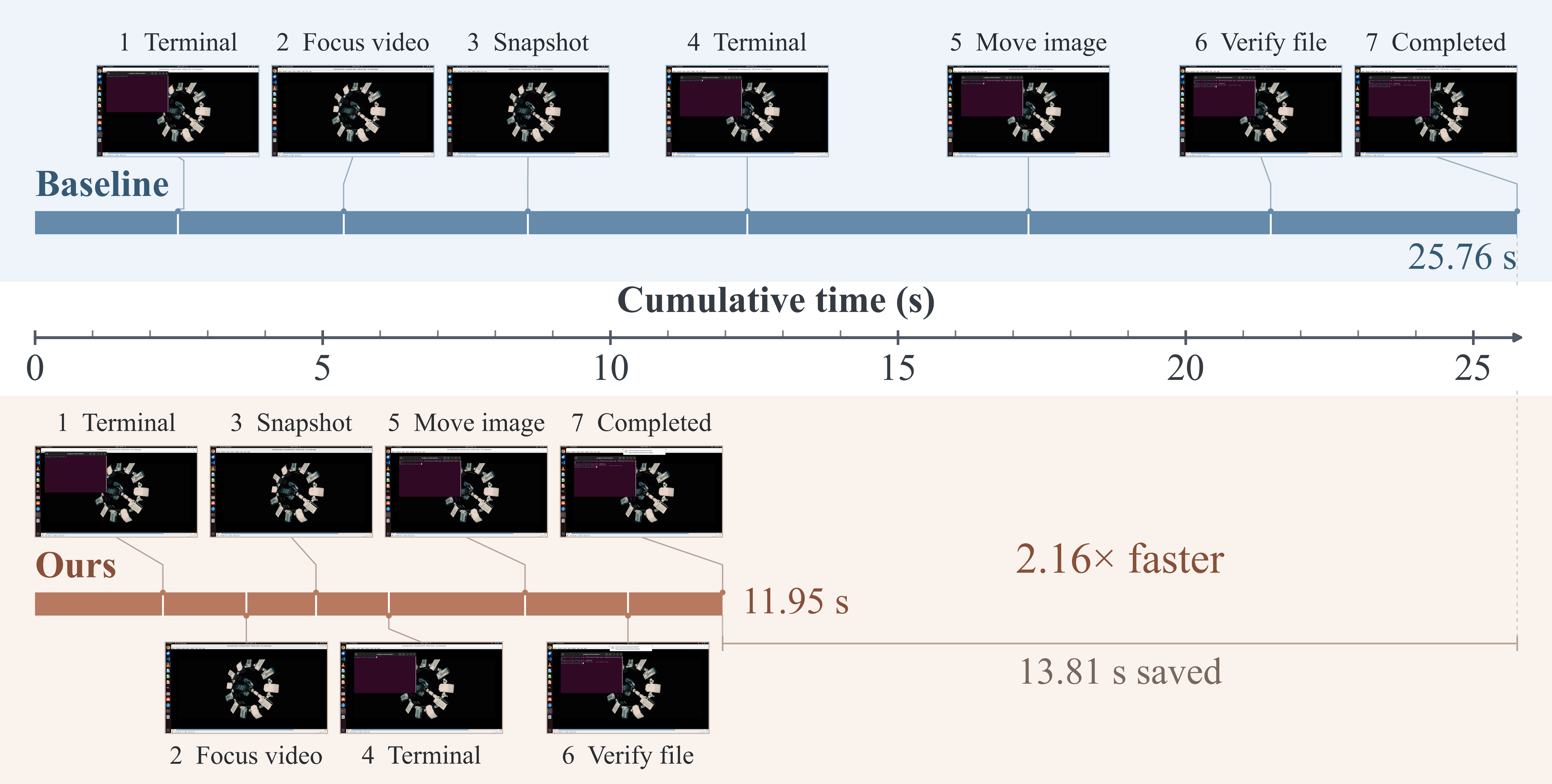}
  \caption{
    \textbf{Qualitative comparison of task execution time on an OSWorld task.}
    Dense inference and \sysname complete the same seven interaction stages.
    \sysname reduces task execution time from $25.76$ to $11.95$ seconds, corresponding to a $2.16\times$ speedup and $13.81$ seconds saved.
    The horizontal axis measures cumulative task execution time.
  }
  \label{fig:qualitative_execution}
\end{figure}

\begin{table}[!t]
  \centering
  \caption{
    \textbf{FLOPs and wall-clock efficiency of \sysname.}
    Total latency covers per-step processing and inference, excluding environment execution.
    Speedups use Dense/\sysname ratios of unrounded means; online records come from different trajectories.
  }
  \label{tab:flops_latency}
  \begingroup
  \footnotesize
  \setlength{\tabcolsep}{3pt}
  \renewcommand{\arraystretch}{1.18}
  \begin{tabularx}{\linewidth}{@{}l l *{4}{>{\centering\arraybackslash}X}@{}}
    \toprule
    \multirow[c]{3}{*}{\textbf{Benchmark}}
    & \multirow[c]{3}{*}{\textbf{Model}}
    & \multirow[c]{3}{*}{\makecell{\textbf{FLOPs}\\(T) $\downarrow$}}
    & \multirow[c]{3}{*}{\makecell{\textbf{Total Latency}\\(ms) $\downarrow$}}
    & \multicolumn{2}{c}{\textbf{Speedup}} \\
    \cmidrule(lr){5-6}
    & & & & \textbf{FLOPs} & \textbf{Latency} \\
    & & & & ($\times$) $\uparrow$ & ($\times$) $\uparrow$ \\
    \midrule
    \multirow{2}{*}{\textbf{AITW}}
    & Qwen3-VL-4B & 9.23 & 470.64 & \textbf{3.46} & \textbf{1.96} \\
    & UI-TARS-1.5-7B & 18.33 & 547.23 & \textbf{3.41} & \textbf{1.94} \\
    \midrule
    \multirow{2}{*}{\textbf{Mind2Web}}
    & Qwen3-VL-4B & 13.37 & 633.91 & \textbf{3.29} & \textbf{1.97} \\
    & UI-TARS-1.5-7B & 25.49 & 603.08 & \textbf{3.37} & \textbf{2.32} \\
    \midrule
    \multirow{2}{*}{\textbf{OSWorld}}
    & Qwen3-VL-8B & 70.85 & 1502.69 & \textbf{3.49} & \textbf{2.51} \\
    & GUI-Owl-1.5-8B & 83.81 & 1713.34 & \textbf{3.32} & \textbf{2.55} \\
    \bottomrule
  \end{tabularx}
  \endgroup
\end{table}

\subsection{Analysis}
\label{sec:analysis}

\textbf{Speedup and Latency Reduction.}
Table~\ref{tab:flops_latency} reports $3.29$--$3.49\times$ speedups across the evaluated offline and online settings, alongside reductions in wall-clock latency by factors of $1.94$--$2.55$.
For example, GUI-Owl's mean processing latency on OSWorld decreases from $4362.91$ to $1713.34$\,ms.
Processing latency also reflects memory access, cache management, and input/output costs, so its reduction differs from the FLOPs reduction.

\textbf{Component Contributions.}
Table~\ref{tab:component_ablation} shows complementary benefits across the cumulative configurations.
Historical KV injection provides the largest initial FLOPs reduction, with sparse visual updates and TokenLease bringing further savings.
With the same TokenLease settings, speculative decoding slightly increases FLOPs but lowers cumulative server processing time from $27.51$ to $17.85$\,s/task.
This is consistent with trading additional verification work for fewer serial decoding calls: state updates reduce repeated computation, while historical drafts improve generation efficiency.

\begin{table}[!t]
  \centering
  \caption{
    \textbf{Cumulative component ablation on Qwen3-VL-8B in OSWorld.}
    Checkmarks indicate enabled components; quality uses 361 tasks, with missing results scored as zero.
    TokenLease uses $L_p=6$ and $M_{\mathrm{cur}}=3$ in both enabled configurations.
    FLOPs and Total average valid task measurements; Total sums server processing time and excludes environment execution.
  }
  \label{tab:component_ablation}
  \begingroup
  \footnotesize
  \setlength{\tabcolsep}{3pt}
  \renewcommand{\arraystretch}{1.15}
  \begin{tabularx}{\linewidth}{@{}c c c c *{4}{>{\centering\arraybackslash}X}@{}}
    \toprule
    \multicolumn{4}{c}{\textbf{Enabled Components}}
    & \multicolumn{2}{c}{\textbf{Task Performance}}
    & \multicolumn{2}{c}{\textbf{Inference Cost}} \\
    \cmidrule(lr){1-4}\cmidrule(lr){5-6}\cmidrule(lr){7-8}
    \makecell{History\\KV}
    & \makecell{Sparse\\Vision}
    & \makecell{Token\\Lease}
    & \makecell{Spec.\\Decode}
    & \makecell{Mean Score\\(\%) $\uparrow$}
    & \makecell{Strict SR\\(\%) $\uparrow$}
    & \makecell{FLOPs\\(T/task) $\downarrow$}
    & \makecell{Total\\(s/task) $\downarrow$} \\
    \midrule
    -- & -- & -- & -- & 24.30 & 23.55 & 2981.19 & 45.43 \\
    $\checkmark$ & -- & -- & -- & 27.33 & 26.32 & 1010.29 & 29.54 \\
    $\checkmark$ & $\checkmark$ & -- & -- & 26.21 & 24.93 & 962.02 & 29.35 \\
    $\checkmark$ & $\checkmark$ & $\checkmark$ & -- & 25.43 & 24.65 & \textbf{819.07} & 27.51 \\
    $\checkmark$ & $\checkmark$ & $\checkmark$ & $\checkmark$ & \textbf{27.90} & \textbf{26.87} & 841.33 & \textbf{17.85} \\
    \bottomrule
  \end{tabularx}
  \endgroup
\end{table}

\begin{table}[!t]
  \centering
  \caption{
    \textbf{Sensitivity to TokenLease probe depth and reuse limit on Qwen3-VL-8B in OSWorld.}
    The layer sweep fixes $M_{\mathrm{cur}}=2$, and the reuse-limit sweep fixes $L_p=6$.
    $\dagger$ marks the shared default value of the swept parameter; quality uses all 361 tasks, with missing results scored as zero.
    Bold marks the best value within each sweep.
  }
  \label{tab:tokenlease_sensitivity}
  \begingroup
  \footnotesize
  \setlength{\tabcolsep}{4pt}
  \renewcommand{\arraystretch}{1.12}
  \begin{tabularx}{\linewidth}{@{}l l *{4}{>{\centering\arraybackslash}X}@{}}
    \toprule
    \textbf{Sweep} & \textbf{Setting}
    & \makecell{\textbf{Mean Score}\\(\%) $\uparrow$}
    & \makecell{\textbf{Strict SR}\\(\%) $\uparrow$}
    & \makecell{\textbf{FLOPs}\\(T/step) $\downarrow$}
    & \makecell{\textbf{Total}\\(s/step) $\downarrow$} \\
    \midrule
    \multirow{4}{*}{\makecell[l]{Probe Depth\\($M_{\mathrm{cur}}=2$)}}
    & $L_p = 3$ & 24.28 & 23.27 & \textbf{224.70} & 3.59 \\
    & $L_p = 6^{\dagger}$ & \textbf{27.35} & \textbf{26.59} & 228.22 & \textbf{3.56} \\
    & $L_p = 9$ & 26.48 & 25.48 & 230.43 & 3.58 \\
    & $L_p = 12$ & 25.25 & 24.65 & 231.01 & 3.62 \\
    \midrule
    \multirow{4}{*}{\makecell[l]{Reuse Limit\\($L_p=6$)}}
    & $M_{\mathrm{cur}} = 1$ & 25.93 & 24.65 & 226.28 & 3.57 \\
    & $M_{\mathrm{cur}} = 2$ & 27.35 & \textbf{26.59} & 228.22 & \textbf{3.56} \\
    & $M_{\mathrm{cur}} = 3^{\dagger}$ & \textbf{27.60} & \textbf{26.59} & 224.76 & 3.57 \\
    & $M_{\mathrm{cur}} = 4$ & 26.76 & 25.48 & \textbf{224.33} & 3.60 \\
    \bottomrule
  \end{tabularx}
  \endgroup
\end{table}

\textbf{Probe Depth and Reuse Limit.}
Table~\ref{tab:tokenlease_sensitivity} characterizes sensitivity to probe depth and reuse limit.
In the layer sweep, $L_p=6$ achieves the highest task scores; shallower probing saves slightly more FLOPs at lower quality, while deeper probing adds cost without improving quality.
At $L_p=6$, $M_{\mathrm{cur}}=3$ gives the highest mean score and ties the best strict success rate; extending reuse to four saves only $0.2\%$ additional FLOPs with lower scores.
The shared defaults, $L_p=6$ and $M_{\mathrm{cur}}=3$, were fixed before the main evaluations; this study analyzes their sensitivity on Qwen3-VL/OSWorld.

\textbf{Qualitative Visualization.}
Figure~\ref{fig:qualitative_execution} compares cumulative task execution time for Dense inference and \sysname on an OSWorld task that both complete through the same interaction stages; the time gap grows over the rollout, reaching a $2.16\times$ speedup.
Appendix~\ref{sec:additional_results} reports split-, category-, and mechanism-level results, Appendix~\ref{sec:appendix_experimental_setup} details the evaluation setup, and Appendix~\ref{sec:additional_qualitative_visualizations} presents eight further successful trajectories with similar time savings.

\section{Conclusion}
\label{sec:conclusion}

We presented GUIFlow, a training-free framework that incrementally updates persistent states across GUI interaction.
By reusing position-aligned historical KV states, selectively refreshing current-observation states, and verifying trajectory-derived drafts, GUIFlow reduces repeated computation across visual encoding, prefill, and decoding.
It achieves up to $3.39\times$ and $3.49\times$ FLOPs speedups on offline benchmarks and OSWorld, respectively, while lowering processing latency and maintaining task performance broadly comparable to dense inference.


\bibliography{reference}
\bibliographystyle{iclr2027_conference}

\clearpage
\appendix
\setcounter{table}{0}
\renewcommand{\thetable}{A\arabic{table}}
\section{Additional Experimental Results}
\label{sec:additional_results}

\subsection{Mind2Web Split-Level Results}
\label{sec:mind2web_split_results}

Table~\ref{tab:mind2web_splits} expands the aggregate Mind2Web results in Table~\ref{tab:offline_navigation} across the Cross-Task, Cross-Website, and Cross-Domain splits while retaining the corresponding speedups.
On Qwen3-VL, GUIFlow achieves a $3.29\times$ speedup with an average Step SR of $49.16\%$, only $0.61$ percentage points below Dense.
This provides a substantially stronger efficiency--accuracy trade-off than VisionZip, which reaches a $1.77\times$ speedup but reduces the average Step SR to $42.09\%$, while the higher-scoring Op-Skip provides only a $1.29\times$ speedup.
On UI-TARS, GUIFlow attains the best Cross-Task result and preserves an average Step SR within $0.27$ points of Dense while delivering a $3.37\times$ speedup, compared with at most $1.54\times$ among the baselines.
Across both backbones, no individual split drops by more than $0.92$ percentage points from Dense.

\begin{table}[t]
  \centering
  \caption{
    \textbf{Split-level results on Mind2Web.}
    Step SR (\%) is macro-averaged within each split, and Avg. is the unweighted mean across the three splits.
    Speedup is the Dense/method mean FLOPs ratio; bold denotes the best result for each model.
  }
  \label{tab:mind2web_splits}
  \begingroup
  \footnotesize
  \setlength{\tabcolsep}{3pt}
  \renewcommand{\arraystretch}{1.12}
  \begin{tabularx}{\linewidth}{@{}l l *{5}{>{\centering\arraybackslash}X}}
    \toprule
    \textbf{Model} & \textbf{Method}
    & \makecell{Cross-Task\\Step SR $\uparrow$}
    & \makecell{Cross-Website\\Step SR $\uparrow$}
    & \makecell{Cross-Domain\\Step SR $\uparrow$}
    & \makecell{Avg.\\Step SR $\uparrow$}
    & \makecell{Speedup\\($\times$) $\uparrow$} \\
    \midrule
    \multirow{8}{*}{\makecell[l]{\textbf{Qwen3-VL}\\\textbf{4B-Instruct}}}
    & Dense & 51.11 & 48.37 & 49.84 & 49.77 & 1.00 \\
    & FastV & 49.55 & 46.73 & 48.64 & 48.31 & 1.46 \\
    & VisionZip & 42.36 & 41.23 & 42.69 & 42.09 & 1.77 \\
    & GUI-KV & 51.59 & 47.06 & 49.09 & 49.25 & 1.00 \\
    & STaR-KV & 51.40 & 47.81 & 49.63 & 49.61 & 1.00 \\
    & Op-Skip & \textbf{54.18} & \textbf{52.48} & \textbf{53.56} & \textbf{53.41} & 1.29 \\
    & HistPrune & 51.09 & 48.22 & 49.93 & 49.75 & 1.27 \\
    & \cellcolor{GUIFlowAccent!15}\sysname
    & \cellcolor{GUIFlowAccent!15}50.59
    & \cellcolor{GUIFlowAccent!15}47.45
    & \cellcolor{GUIFlowAccent!15}49.45
    & \cellcolor{GUIFlowAccent!15}49.16
    & \cellcolor{GUIFlowAccent!15}\textbf{3.29} \\
    \midrule
    \multirow{8}{*}{\makecell[l]{\textbf{UI-TARS}\\\textbf{1.5-7B}}}
    & Dense & 55.04 & 52.19 & \textbf{52.98} & \textbf{53.40} & 1.00 \\
    & FastV & 54.46 & 51.59 & 51.14 & 52.40 & 1.54 \\
    & VisionZip & 51.98 & 48.08 & 47.87 & 49.31 & 1.49 \\
    & GUI-KV & 53.30 & 51.99 & 51.21 & 52.17 & 1.00 \\
    & STaR-KV & 54.51 & 51.88 & 52.76 & 53.05 & 1.00 \\
    & Op-Skip & 51.16 & 47.69 & 49.62 & 49.49 & 1.24 \\
    & HistPrune & 54.74 & \textbf{52.22} & 52.91 & 53.29 & 1.20 \\
    & \cellcolor{GUIFlowAccent!15}\sysname
    & \cellcolor{GUIFlowAccent!15}\textbf{55.13}
    & \cellcolor{GUIFlowAccent!15}51.65
    & \cellcolor{GUIFlowAccent!15}52.62
    & \cellcolor{GUIFlowAccent!15}53.13
    & \cellcolor{GUIFlowAccent!15}\textbf{3.37} \\
    \bottomrule
  \end{tabularx}
  \endgroup
\end{table}

\subsection{OSWorld Category-Level Results}
\label{sec:osworld_category_results}

Table~\ref{tab:osworld_categories} reports category-level efficiency and quality results for GUI-Owl on OSWorld.
Overall, GUIFlow reduces FLOPs per request by $69.9\%$ ($3.32\times$) and total FLOPs per task by $67.7\%$ ($3.10\times$), while keeping the mean score within $2.49$ points of Dense.
Task-level speedups range from $2.33\times$ to $4.41\times$ across application categories, demonstrating consistent computational savings under varying request counts.

\begin{table}[t]
  \centering
  \caption{
    \textbf{Category-level efficiency and quality on OSWorld with GUI-Owl-1.5-8B.}
    Request and score changes are measured relative to Dense; FLOPs speedups are computed as Dense/\sysname.
    Overall covers all 361 tasks; efficiency uses valid measurements.
  }
  \label{tab:osworld_categories}
  \begingroup
  \footnotesize
  \setlength{\tabcolsep}{4pt}
  \renewcommand{\arraystretch}{1.12}
  \begin{tabular*}{\linewidth}{@{\extracolsep{\fill}}l r r r r r@{}}
    \toprule
    \textbf{Application} & \textbf{Tasks} & \makecell{\textbf{Request}\\\textbf{change (\%)}} & \makecell{\textbf{FLOPs/request}\\\textbf{speedup ($\times$)}} & \makecell{\textbf{Task FLOPs}\\\textbf{speedup ($\times$)}} & \makecell{\textbf{Score change}\\\textbf{(pp)}} \\
    \midrule
    Chrome & 46 & $+40.4$ & 3.27 & 2.33 & $-0.01$ \\
    GIMP & 26 & $+14.2$ & 3.44 & 3.01 & $-3.85$ \\
    LibreOffice Calc & 47 & $-5.6$ & 3.23 & 3.42 & $-6.38$ \\
    LibreOffice Impress & 47 & $-16.8$ & 3.67 & 4.41 & $+4.26$ \\
    LibreOffice Writer & 23 & $+45.5$ & 3.45 & 2.37 & $-8.70$ \\
    Multi-Apps & 93 & $+5.7$ & 3.19 & 3.02 & $-1.08$ \\
    OS & 24 & $+16.8$ & 3.33 & 2.86 & $0.00$ \\
    Thunderbird & 15 & $-15.0$ & 3.64 & 4.28 & $-6.67$ \\
    VLC & 17 & $+8.8$ & 3.54 & 3.25 & $-11.76$ \\
    VS Code & 23 & $-6.9$ & 3.53 & 3.80 & $-4.35$ \\
    \midrule
    \textbf{Overall} & \textbf{361} & \textbf{$+7.1$} & \textbf{3.32} & \textbf{3.10} & \textbf{$-2.49$} \\
    \bottomrule
  \end{tabular*}
  \endgroup
\end{table}

\subsection{TokenLease Mechanistic Statistics}
\label{sec:tokenlease_statistics}

Table~\ref{tab:tokenlease_statistics} complements the outcome-level sensitivity results in Table~\ref{tab:tokenlease_sensitivity} by exposing the internal reuse behavior of TokenLease.
Across probe depths, the acceptance rate remains stable at approximately $91.6\%$, while moving the probe from layer 3 to layer 12 reduces the fraction of reused image tokens from $30.66\%$ to $23.64\%$ and the FLOPs reduction from $9.07\%$ to $6.23\%$.
This indicates that deeper probing primarily reduces the amount of downstream computation available for reuse rather than materially changing the lease decisions.
With $L_p=6$, increasing $M_{\mathrm{cur}}$ from 1 to 3 raises image-token reuse from $25.60\%$ to $31.31\%$ and the FLOPs reduction from $8.08\%$ to $8.77\%$.
Extending the limit to 4 yields no further computational benefit and slightly lowers the paired latency speedup.
These statistics complement Table~\ref{tab:tokenlease_sensitivity} by characterizing reuse behavior around the shared defaults, $L_p=6$ and $M_{\mathrm{cur}}=3$.

\begin{table}[t]
  \centering
  \caption{
    \textbf{Mechanistic statistics of TokenLease on Qwen3-VL-8B in OSWorld.}
    The probe-depth sweep fixes $M_{\mathrm{cur}}=2$, and the reuse-limit sweep fixes $L_p=6$.
    FLOPs reduction and latency speedup are paired against Dense on common task-step keys; $\dagger$ marks the shared default value of the swept parameter.
  }
  \label{tab:tokenlease_statistics}
  \begingroup
  \footnotesize
  \setlength{\tabcolsep}{3pt}
  \renewcommand{\arraystretch}{1.12}
  \begin{tabularx}{\linewidth}{@{}l l c *{4}{>{\centering\arraybackslash}X}@{}}
    \toprule
    \textbf{Sweep} & \textbf{Setting} & \makecell{Effective\\Tasks}
    & \makecell{Lease\\Acceptance (\%)}
    & \makecell{Reused Image\\Tokens (\%)}
    & \makecell{FLOPs\\Reduction (\%)}
    & \makecell{Latency\\Speedup ($\times$)} \\
    \midrule
    \multirow{4}{*}{\makecell[l]{Probe Depth\\($M_{\mathrm{cur}}=2$)}}
    & $L_p = 3$ & 359 & 91.66 & 30.66 & 9.07 & 1.05 \\
    & $L_p = 6^{\dagger}$ & 359 & 91.59 & 29.95 & 7.53 & 1.06 \\
    & $L_p = 9$ & 359 & 91.66 & 25.46 & 6.73 & 1.05 \\
    & $L_p = 12$ & 359 & 91.56 & 23.64 & 6.23 & 1.04 \\
    \midrule
    \multirow{4}{*}{\makecell[l]{Reuse Limit\\($L_p=6$)}}
    & $M_{\mathrm{cur}} = 1$ & 359 & 91.54 & 25.60 & 8.08 & 1.05 \\
    & $M_{\mathrm{cur}} = 2$ & 359 & 91.59 & 29.95 & 7.53 & 1.06 \\
    & $M_{\mathrm{cur}} = 3^{\dagger}$ & 358 & 91.58 & 31.31 & 8.77 & 1.05 \\
    & $M_{\mathrm{cur}} = 4$ & 359 & 91.43 & 30.86 & 8.73 & 1.04 \\
    \bottomrule
  \end{tabularx}
  \endgroup
\end{table}

\newpage
\section{Experimental Setup}
\label{sec:appendix_experimental_setup}

\textbf{Benchmarks and Metrics.}
We evaluate offline navigation on AITW~\citep{rawles2023androidinthewild} and the screenshot-based Mind2Web benchmark~\citep{deng2023mind2web,cheng2024seeclick}.
AITW contains five subsets---General, Install, GoogleApps, Single, and WebShopping---comprising 4,694 evaluated steps in our setup.
We follow the official action-matching metric and report the macro-average over the five subsets; action types must match, taps are evaluated against the annotated target region, and swipes must agree along the primary direction.
For Mind2Web, we evaluate 9,268 runnable steps from 1,341 tasks under teacher-forced offline navigation, providing up to four preceding screenshot--action pairs together with the current screenshot.
A step is successful when the predicted point lies within the ground-truth element and the predicted operation matches the reference; we macro-average within Cross-Task, Cross-Website, and Cross-Domain and report their unweighted mean.
For online evaluation, we use all 361 tasks in OSWorld's \texttt{test\_nogdrive} split~\citep{xie2024osworld}.
The agent interacts with the environment in closed loop, and task quality is measured by the environment-specific final-state evaluator.
Mean Score and strict success rate use all 361 tasks as the denominator; only an evaluator score of one counts as strict success.
Missing or incomplete results, including tasks interrupted by simulator crashes, are assigned zero scores and counted as failures.

\textbf{Models.}
Our evaluation covers both general-purpose vision-language models and GUI-specialized agents.
For offline navigation, we use Qwen3-VL-4B-Instruct~\citep{bai2025qwen3} and UI-TARS-1.5-7B, from the UI-TARS family~\citep{qin2025ui}.
For online OSWorld interaction, we use Qwen3-VL-8B-Instruct and GUI-Owl-1.5-8B-Instruct~\citep{xu2026mobile}.
This selection evaluates GUIFlow across two model scales and across both general multimodal and GUI-native model families.
GUIFlow is applied at inference time without updating model parameters, and each model retains its native processor, prompt structure, and action representation; UI-TARS uses its \texttt{start\_box} action protocol in the offline evaluation.

\textbf{Hyperparameter Configuration.}
We determined a shared hyperparameter configuration through preliminary observations on sampled offline tasks before the main evaluations.
The configuration, including $L_p=6$ and $M_{\mathrm{cur}}=3$, was then fixed across all evaluated models and both offline and online benchmarks, except for the explicitly varied settings in the ablation and sensitivity studies.
The OSWorld sensitivity study analyzes parameter effects and was not used to select the defaults.

\textbf{Baselines and Efficiency Measurement.}
We compare against Dense inference, visual-token pruning, KV compression, operator skipping, and historical caching baselines.
All experiments are conducted on NVIDIA GeForce RTX 4090 GPUs.
We measure efficiency using average VLM FLOPs and processing latency, and report improvements relative to Dense inference.
Efficiency statistics retain valid measurements from successful and failed tasks; missing measurements from simulator crashes are excluded rather than imputed as zeros.

\clearpage
\section{Additional Qualitative Visualizations}
\label{sec:additional_qualitative_visualizations}

\begin{figure}[H]
  \centering
  \begin{subfigure}[t]{\linewidth}
    \centering
    \includegraphics[width=\linewidth]{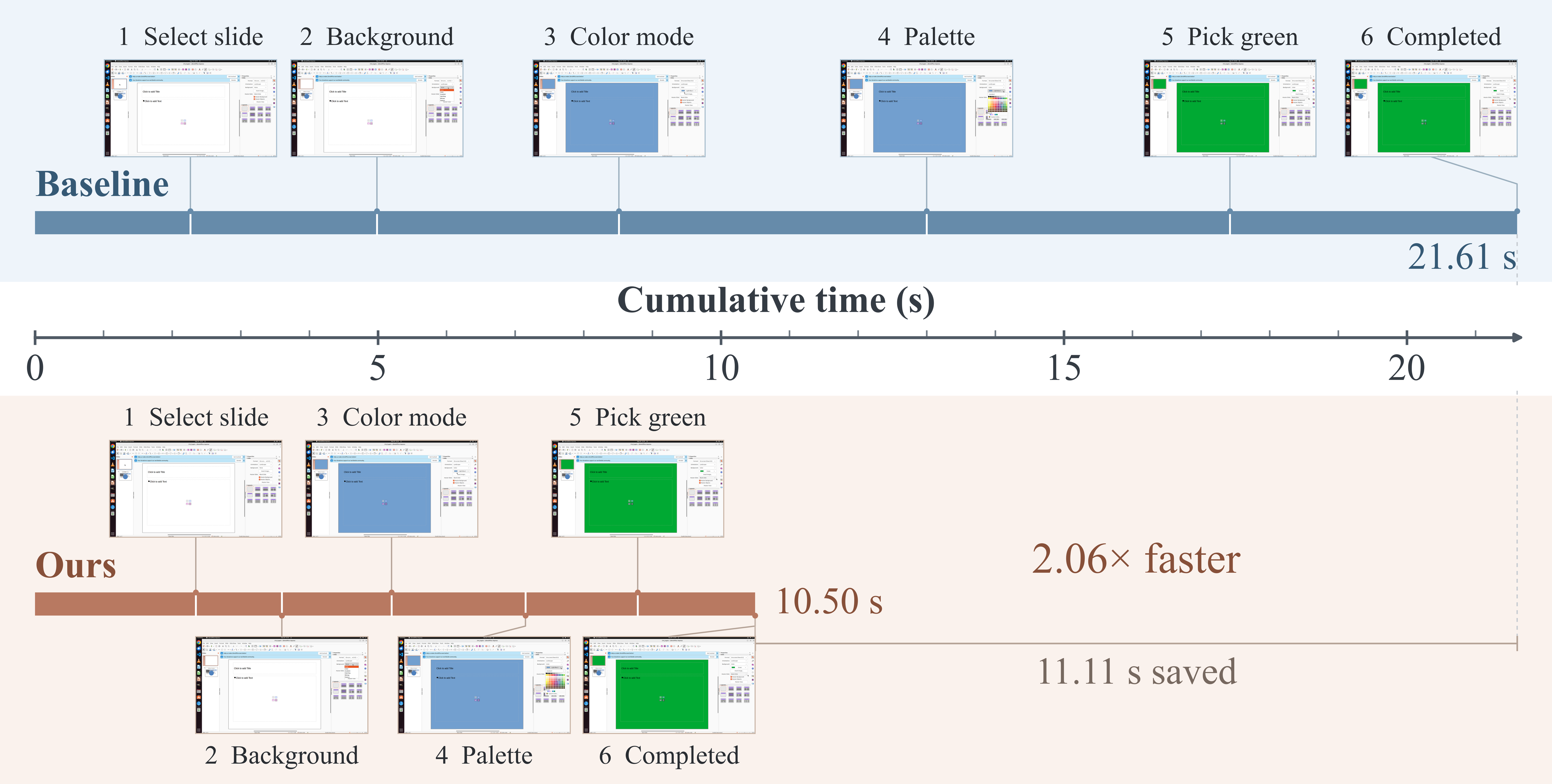}
    \caption{Changing a slide background to green: $21.61$ to $10.50$ s ($2.06\times$).}
  \end{subfigure}\\[0.6em]
  \begin{subfigure}[t]{\linewidth}
    \centering
    \includegraphics[width=\linewidth]{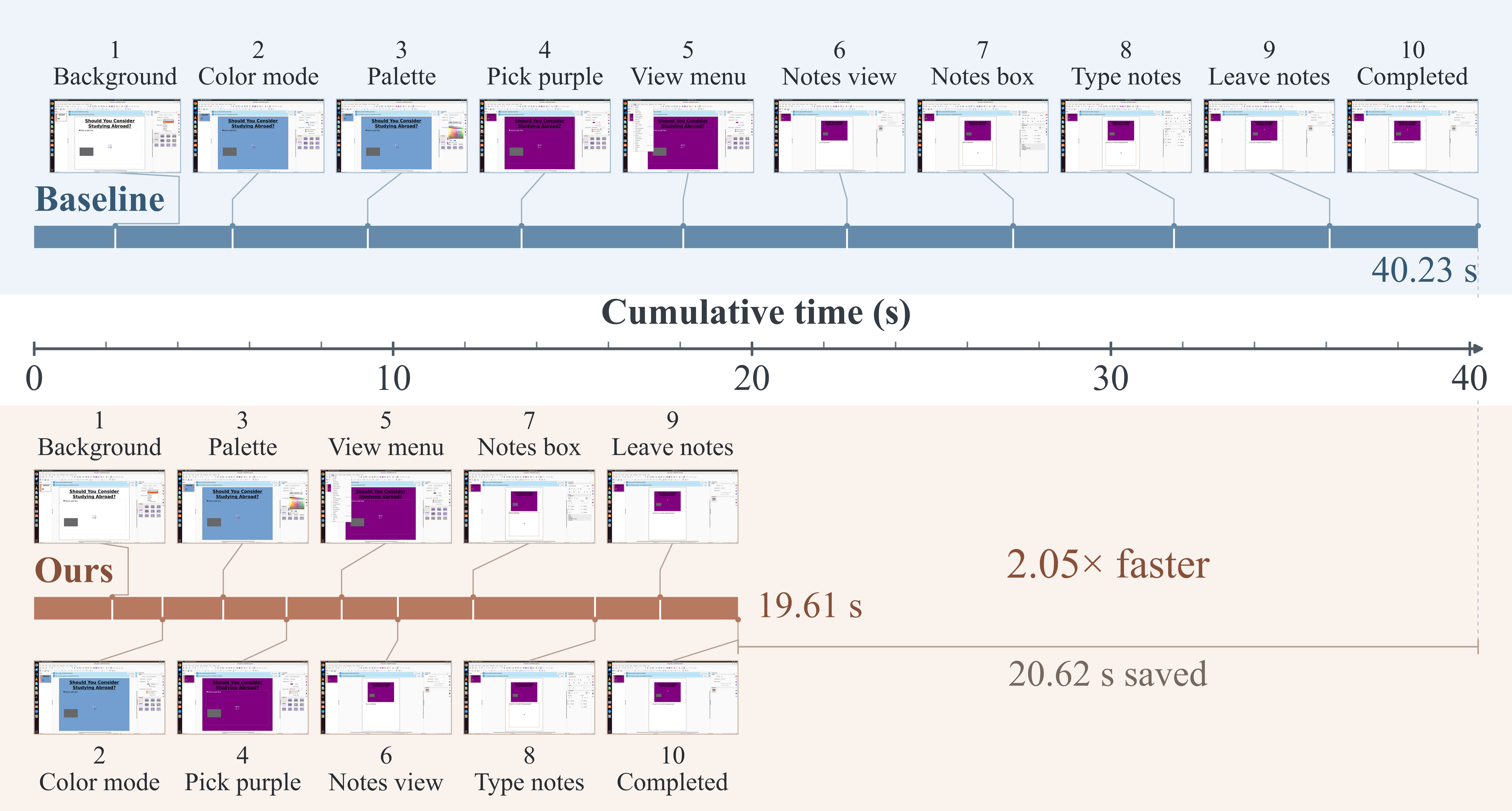}
    \caption{Editing slide color and speaker notes: $40.23$ to $19.61$ s ($2.05\times$).}
  \end{subfigure}
  \caption{\textbf{Additional presentation-editing trajectories.} \sysname preserves all interaction stages while reducing cumulative task execution time by $11.11$ and $20.62$ seconds, respectively.}
  \label{fig:qualitative_impress}
\end{figure}

\begin{figure}[t]
  \centering
  \begin{subfigure}[t]{\linewidth}
    \centering
    \includegraphics[width=\linewidth]{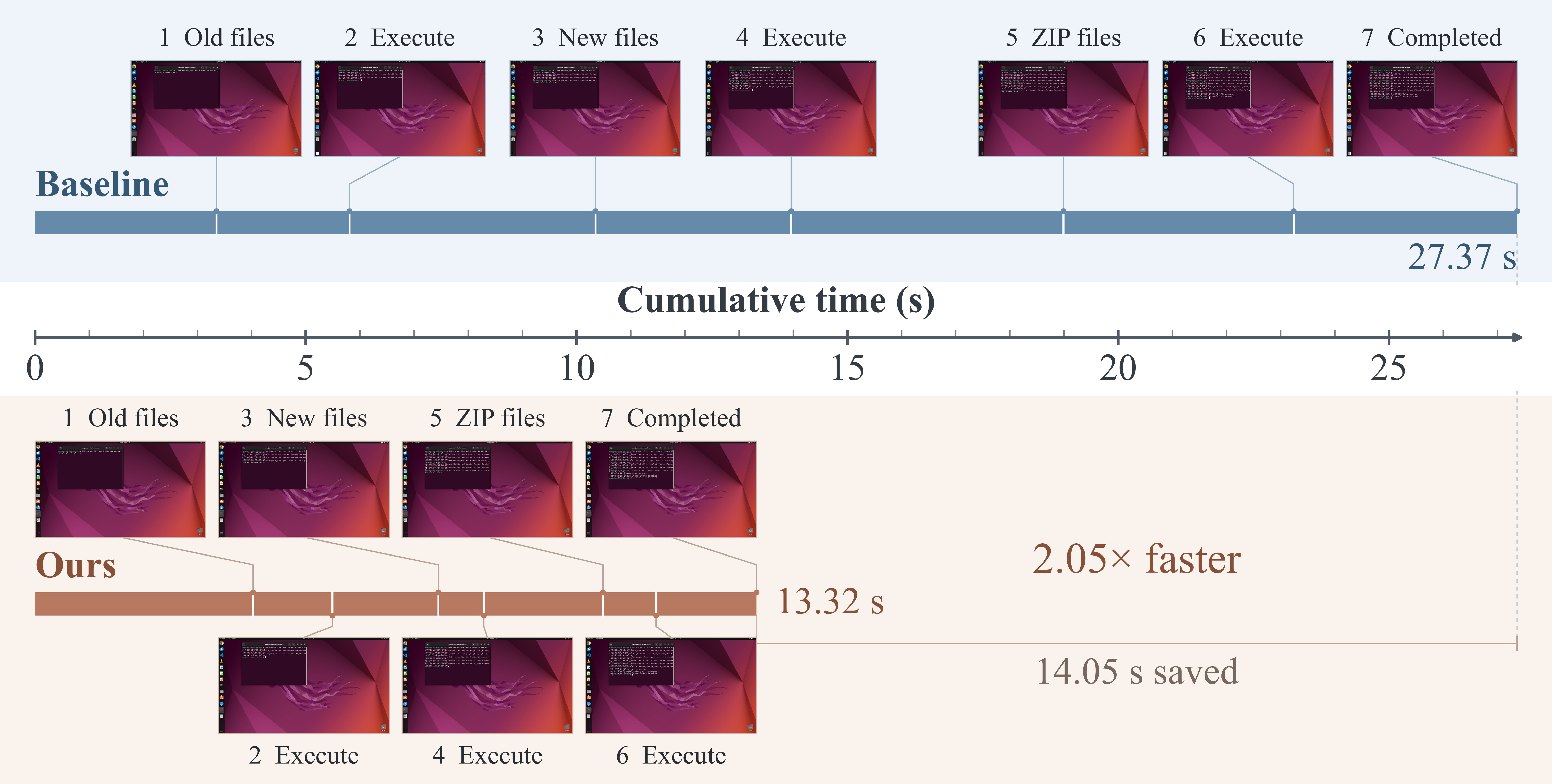}
    \caption{Sorting and archiving files: $27.37$ to $13.32$ s ($2.05\times$).}
  \end{subfigure}\\[0.6em]
  \begin{subfigure}[t]{\linewidth}
    \centering
    \includegraphics[width=\linewidth]{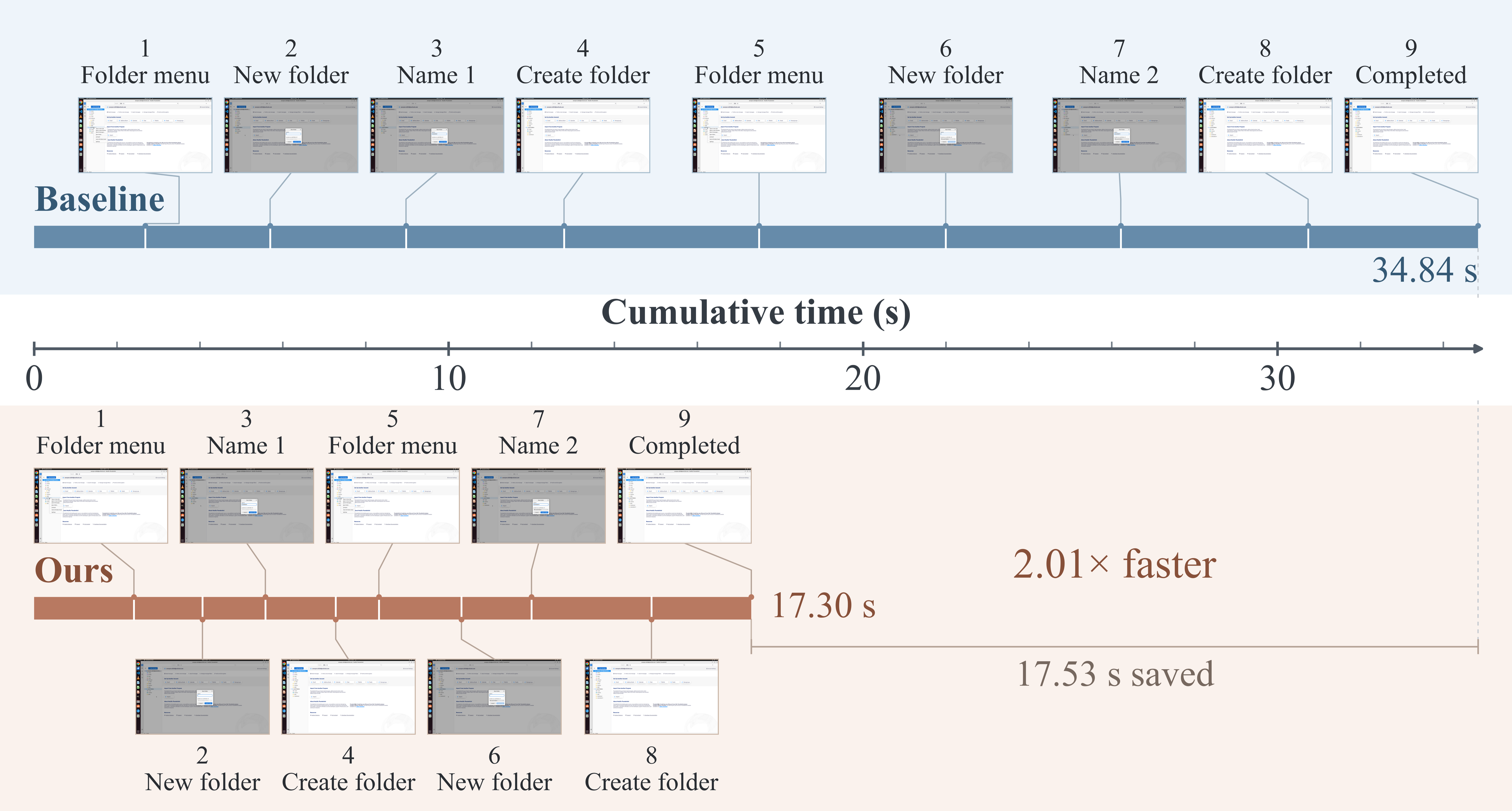}
    \caption{Creating two mail folders: $34.84$ to $17.30$ s ($2.01\times$).}
  \end{subfigure}
  \caption{\textbf{Additional file- and email-organization trajectories.} \sysname saves $14.05$ and $17.53$ seconds while retaining the completed task sequence.}
  \label{fig:qualitative_files}
\end{figure}

\begin{figure}[t]
  \centering
  \begin{subfigure}[t]{\linewidth}
    \centering
    \includegraphics[width=\linewidth]{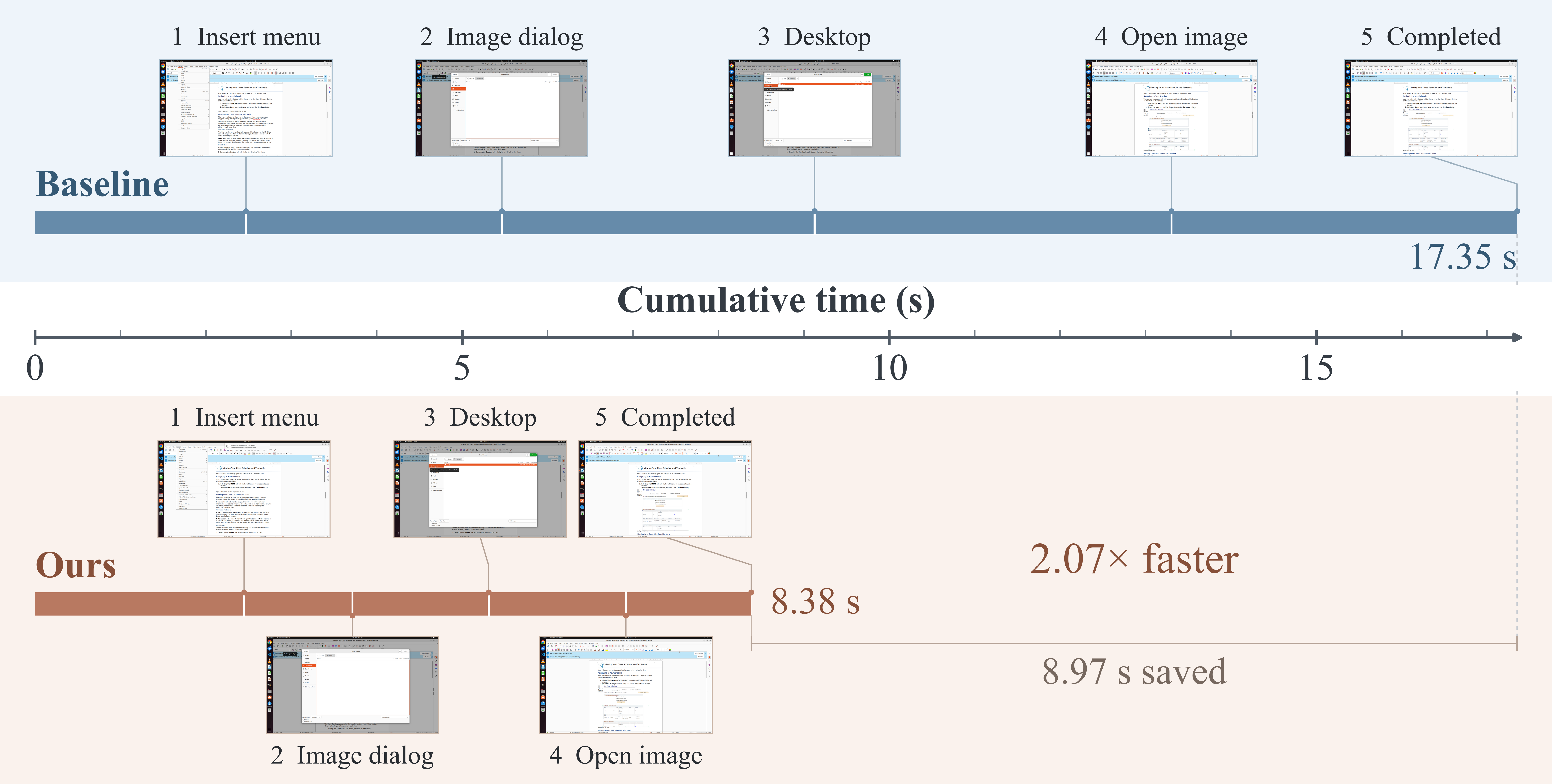}
    \caption{Inserting an image in Writer: $17.35$ to $8.38$ s ($2.07\times$).}
  \end{subfigure}\\[0.6em]
  \begin{subfigure}[t]{\linewidth}
    \centering
    \includegraphics[width=\linewidth]{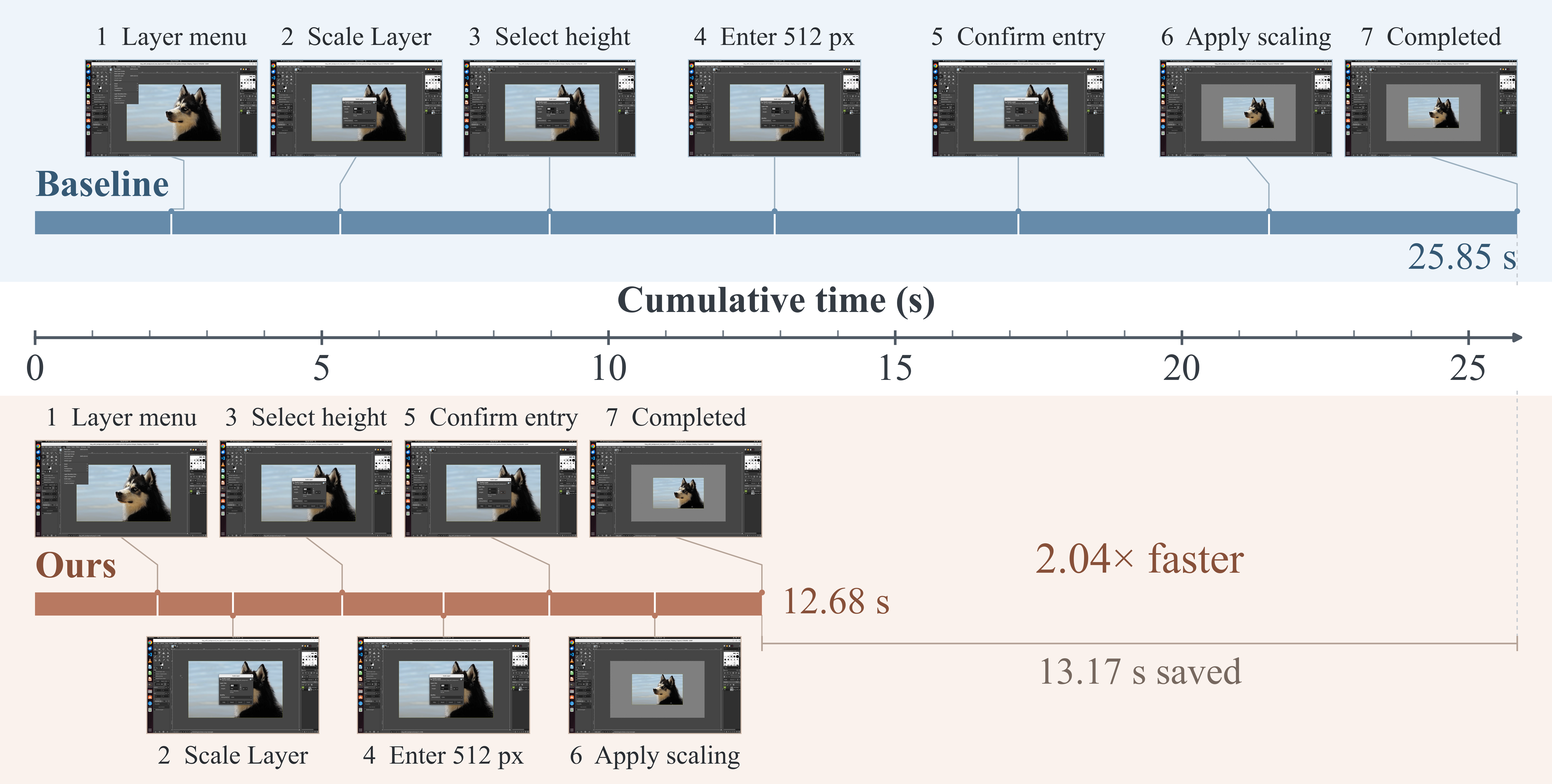}
    \caption{Resizing an image in GIMP: $25.85$ to $12.68$ s ($2.04\times$).}
  \end{subfigure}
  \caption{\textbf{Additional document- and image-editing trajectories.} \sysname saves $8.97$ and $13.17$ seconds while completing the same editing stages as Dense inference.}
  \label{fig:qualitative_editing}
\end{figure}

\begin{figure}[t]
  \centering
  \begin{subfigure}[t]{\linewidth}
    \centering
    \includegraphics[width=\linewidth]{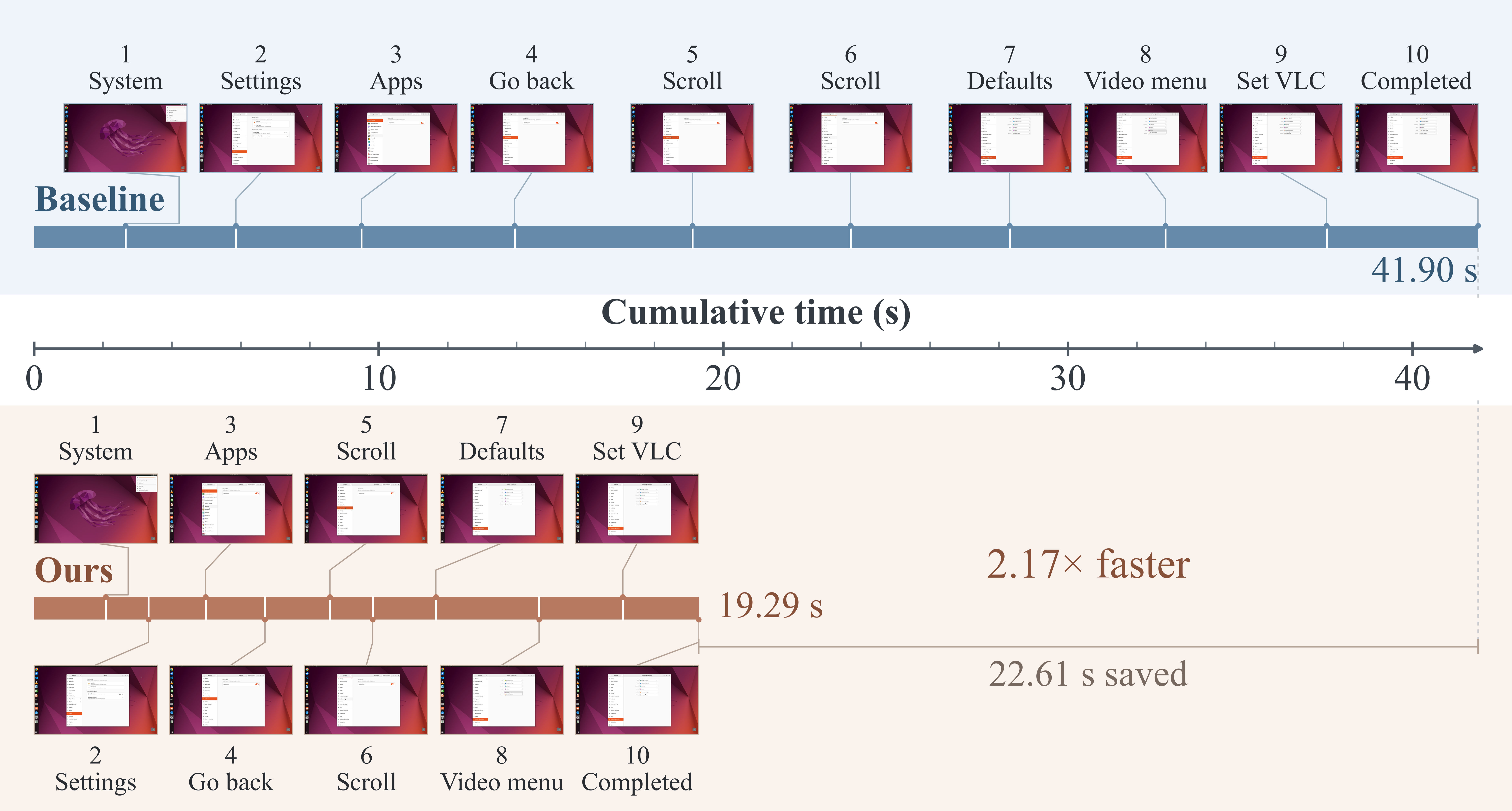}
    \caption{Setting VLC as the default application: $41.90$ to $19.29$ s ($2.17\times$).}
  \end{subfigure}\\[0.6em]
  \begin{subfigure}[t]{\linewidth}
    \centering
    \includegraphics[width=\linewidth]{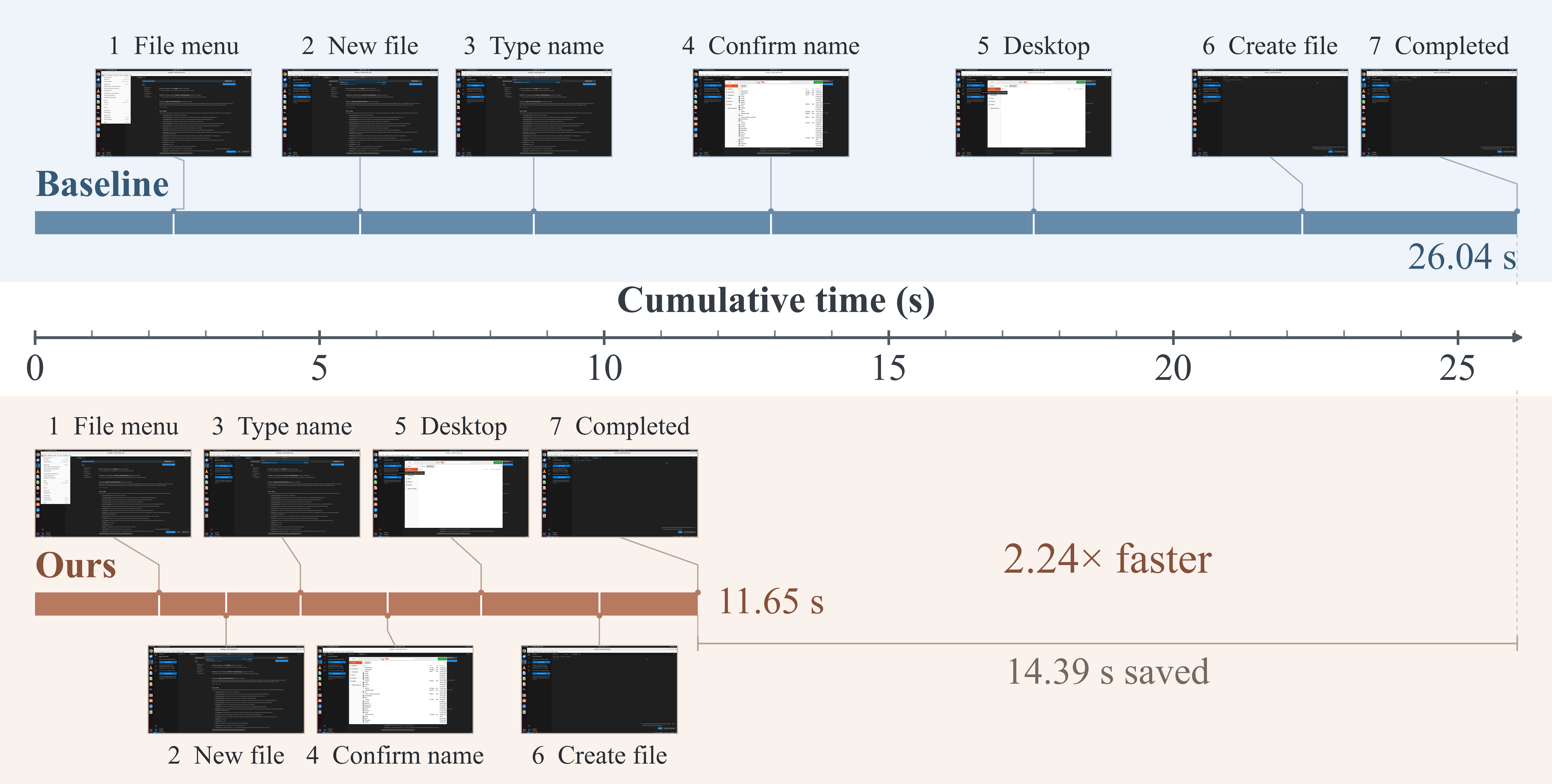}
    \caption{Creating a Python file in VS Code: $26.04$ to $11.65$ s ($2.24\times$).}
  \end{subfigure}
  \caption{\textbf{Additional system-configuration and code-editing trajectories.} \sysname saves $22.61$ and $14.39$ seconds, respectively, while preserving all interaction stages.}
  \label{fig:qualitative_system}
\end{figure}

Figures~\ref{fig:qualitative_impress}--\ref{fig:qualitative_system} provide eight additional successful OSWorld trajectories spanning presentation editing, file operations, document editing, image manipulation, email organization, system configuration, and code editing.
In every example, Dense inference and \sysname follow the same interaction stages and reach the completed state, while \sysname consistently accumulates less task execution time.
Across these trajectories, the observed speedups range from $2.01\times$ to $2.24\times$, with $8.97$--$22.61$ seconds saved per task.
These trajectories complement Figure~\ref{fig:qualitative_execution} by illustrating lower cumulative task execution time across applications and trajectory lengths.
All timelines report task execution time.

\end{document}